\documentclass{article}
\usepackage[a4paper,left=1in,right=1in,top=3cm,bottom=2cm]{geometry}

\usepackage{graphicx}
\usepackage{amsmath}
\usepackage{algorithmic}
\usepackage{algorithm}

\usepackage{booktabs} 
\usepackage{array}

\usepackage[para,online,flushleft]{threeparttable}
\usepackage{diagbox}

\begin{document}
\title{Co-Learning Bayesian Optimization}

\author{Zhendong~Guo,
        Yew-Soon~Ong,~\textit{Fellow,~IEEE,}
        ~Tiantian~He,
        and Haitao~Liu
\thanks{Zhendong Guo and Tiantian He are with the Data Science and Artificial Intelligence Research Center, School of Computer Science and Engineering, Nanyang Technological University, Singapore 639798. (e-mail: \{zhendong.guo, tiantian.he\}@ntu.edu.sg).}
\thanks{Yew-Soon Ong is with School of Computer Science and Engineering, Nanyang Technological University, Singapore 639798 (e-mail: ASYSONG@ntu.edu.sg).}
\thanks{Haitao Liu is with the School of Energy and Power Engineering, Dalian University of Technology,116024, Dalian, China (e-mail: htliu@dlut.edu.cn).}
}


\date{}
\maketitle

\begin{abstract}
Bayesian optimization (BO) is well known to be sample-efficient for solving black-box problems. 
However, the BO algorithms can sometimes get stuck in suboptimal solutions even with plenty of samples. 
Intrinsically, such suboptimal problem of BO can attribute to the poor surrogate accuracy of the trained Gaussian process (GP), particularly that in the regions where the optimal solutions locate.
Hence, we propose to build multiple GP models instead of a single GP surrogate to complement each other and thus resolving the suboptimal problem of BO.
Nevertheless, according to the bias-variance tradeoff equation, the individual prediction errors can increase when increasing the diversity of models, which may lead to even worse overall surrogate accuracy.  
On the other hand, based on the theory of Rademacher complexity, it has been proved that exploiting the agreement of models on unlabeled information can help to reduce the complexity of the hypothesis space, and therefore achieving the required surrogate accuracy with fewer samples. Such value of model agreement has been extensively demonstrated for co-training style algorithms to boost model accuracy with a small portion of samples.
Inspired by the above, we propose a novel BO algorithm labeled as co-learning BO (CLBO), which exploits both model diversity and agreement on unlabeled information to improve the overall surrogate accuracy with limited samples, and therefore achieving more efficient global optimization. 
Through tests on five numerical toy problems and three engineering benchmarks, the effectiveness of proposed CLBO has been well demonstrated.

\end{abstract}

\textbf{Index Terms:} Bayesian optimization, co-training style semi-supervised learning, black-box problems, robotics control, aircraft design.


\section{Introduction}


Bayesian optimization is well-known to be sample efficient, which has drawn wide attentions for solving black-box problems~\cite{martinez2018funneled},~\cite{2021Guo},~\cite{song2016research}, such as active control of robotics systems~\cite{martinez2017feeling} and hyperparameter tuning~\cite{swersky2013multi}.
The basic process of BO is as follows.
First, it uses Gaussian process (GP)~\cite{rasmussen2010gaussian} to build a surrogate of the underlying objective function.
Second, it employs an acquisition function~\cite{jones1998efficient},~\cite{couckuyt2014fast},~\cite{hennig2012entropy},~\cite{wu2016parallel}
which makes use of the information extracted from GP to select the most promising solution candidate to query. 
The algorithm repeats the above process until the stopping condition is met.

Though tremendous success has been reported~\cite{stuckman1992comparison},~\cite{huang2006global}, BO approaches sometimes can get stuck in poor solutions even with plenty of adaptively and sequentially selected samples~\cite{bull2011convergence},~\cite{qin2017improving},~\cite{2002Flexibility}.
\emph{Intrinsically, such suboptimal problem of BO can attribute to the poor accuracy of the trained GP surrogate, particularly that in the region where the true optima locate}~\cite{shahriari2015taking}.
To improve the accuracy of GP surrogate in BO process, some studies resort to extracting information from the related source tasks~\cite{joy2019flexible},~\cite{min2020generalizing} or building multi-fidelity surrogate~\cite{kandasamy2017multi},~\cite{guo2018analysis},~\cite{2021Parallel}.
Nevertheless, neither the source task nor multi-fidelity samples can be always available. Alternatively, as no single model can be most accurate for all the scenarios~\cite{gupta2017insights},~\cite{zhang2015multimodel},~\cite{min2017multiproblem},~\cite{zhou2006combining}, some other studies propose to use multiple models~\cite{martinez2018funneled},~\cite{2013Efficient} instead of a single GP to improve the surrogate accuracy in BO process.

Note that, the idea of using multiple models has been extensively explored in the machine learning community~\cite{he2019contextual},~\cite{xu2013survey}, such as ensemble learning~\cite{zhou2009semi}.
In particular, the overall accuracy by using multiple models at a single point site or over a region can be calculated by the bias-variance tradeoff equation~\cite{krogh1995neural}:
\begin{equation}
\begin{small}
\begin{aligned}
\underbrace {{{({{\hat y}_{}} - y)}^2}}_E &= \underbrace {\sum {{w_i}{{\left( {{{\hat y}^{(i)}} - y} \right)}^2}} }_{{E_{ind}}} - \underbrace {\sum {{w_i}{{\left( {{{\hat y}^{(i)}} - {{\hat y}_{}}} \right)}^2}} }_{Diversity}\\
{{\hat y}_{}} &= \sum\limits_i {{w_i}{{\hat y}^{(i)}},\;} \sum\limits_i {{w_i}}  = 1
\end{aligned}
\end{small}
\end{equation}
where, ${{\hat y}^{(i)}}$ and ${\hat y_{}}$ denote the overall prediction and the individual model prediction, respectively, and ${w_i}$ is the corresponding weight. 
The term \begin{small}${({\hat y_{}} - y)^2}$\end{small} is the error of the overall model, denoted by $E$; \begin{small}$\sum {{w_i}{{({{\hat y}^{(i)}} - y)}^2}}$\end{small} is the error of the individual models, denoted by ${E_{ind}}$; and \begin{small}$\sum {{w_i}{{({{\hat y}^{(i)}} - {{\hat y}_{}})}^2}}$\end{small} measures the diversity among individual models, which is also known as the term of ambiguity.
Ideally, simultaneously increasing the model diversity and reducing the error of individual models (i.e.,$E_{ind}$) can reduce the overall prediction error.
However, \emph{purely increasing the diversity among models can lead to the increase of $E_{ind}$, which may result in even worse overall prediction error}.
Hence, \emph{both the individual model accuracy and model diversity should be taken into account when building multiple models}.

On the other hand, according to the theory of Rademacher complexity~\cite{Luca2015Local}, \emph{the bound of model prediction loss (or error) is related to the complexity of the hypothesis searching space}~\cite{leskes2008value}:
\begin{equation}
\begin{small}
\begin{aligned}
E\left( {L(Y,f(X))} \right) &\le {{\hat E}_n}\left( {L(Y,f(X))} \right)\\
&+ \underbrace {{R_n}(\tilde L \circ F)}_{Complexity} + \sqrt {\frac{{8\log (2/\delta )}}{n}}
\end{aligned}
\end{small}
\end{equation}
where, \begin{small}$E\left( {L(Y,f(X))} \right)$\end{small} and \begin{small}${{\hat E}_n}\left( {L(Y,f(X))} \right)$\end{small} are the expectations of the real loss and empirical loss, respectively, \begin{small}${R_n}(\tilde L \circ F)$\end{small} is the Rademacher complexity of the hypothesis space, and $n$ and $\delta $ are the number of training samples and the certainty in the success of the whole procedure, respectively.

As proved in~\cite{leskes2008value},~\cite{farquhar2006two}, \emph{the agreement of models on unlabeled information can reduce the complexity of hypothesis space (i.e.,\begin{small}${R_n}(\tilde L \circ F)$\end{small} in Eq.(2)), and therefore helping to reduce the number of samples necessary to achieve required model accuracy}. 
The value of model agreement as shown as above has been extensively demonstrated by co-training style algorithms~\cite{xu2013survey},~\cite{blum1998combining},~\cite{navaratnam2007joint}, which has achieved tremendous success in boosting the model accuracy with limited available samples~\cite{zhou2004democratic},~\cite{muslea2006active}.


Inspired by the above, we propose a novel BO algorithm labeled as co-learning BO (CLBO).
The motivation behind CLBO is as follows.
First, since no surrogate model can be most accurate at all sites of the design space, we propose to build multiple GP models to improve the overall surrogate accuracy, particularly that in the region where the true optima locate.  
Second, based on the analysis with Eq.(2), we propose to enforce model agreement on unlabeled information to
reduce the sample complexity of hypothesis space, and therefore building reasonably accurate individual models with limited training samples.
And further, these reasonably accurate individual models are combined, and the differences between these individual models can help them to complement each other. Following the way shown as above, the suboptimal problem of BO can be addressed by much better overall surrogate accuracy particularly that in the vicinity of real optima, and therefore better optimization results can be expected.

With the above art in mind, we use two kinds of GP models in CLBO.
On one hand, we use multi-output GP (MOGP)~\cite{bonilla2008multi} with the subsets of training samples to obtain multiple GP predictions, wherein we treat the predictions as ``subtasks" and enforce model agreement on curve bumpiness to ensure reasonable individual model accuracy.
On the other hand, though it cannot be most accurate at every site of the design space, the model trained with more samples can
have relatively better mean accuracy over space, therefore we also build a single-output GP surrogate with the full training set, labeled as SOGP.
Then, the combination of MOGP and SOGP models can be expected to have good surrogate accuracy over space, which helps to achieve the optimal solutions more efficiently.


The main contribution of this work can be summarized as follows:

(1) Inspired by the bias-variance tradeoff equation and agreement-assisted co-training paradigm, we propose a novel BO framework with multiple GP models, labeled as co-learning BO (CLBO).
In particular, we propose to exploit both
model diversity and agreement on unlabeled information to improve the overall surrogate accuracy in BO process, and therefore achieving more efficient global optimization.

(2) In doing so, we use the structure of multi-output GP to enforce model agreement on curve bumpiness to obtain individual predictions of good accuracy. 
In the meantime, we build one more single-output GP with full samples to further improve the overall surrogate accuracy in BO process.

(3) Through tests on benchmark functions and engineering problems, and comparison against other BO algorithms, the effectiveness of CLBO is well demonstrated.

The remainder of this paper is organized as follows: Section II presents the preliminaries and related work of CLBO. And then, Section III illustrates the details of proposed CLBO. After that, Section IV shows the experimental studies on the benchmark functions and engineering problems. And finally, we draw conclusions in Section V.

\section{Background and related work}

In this section, the basics of BO are introduced, therein we show the the motivation of using multiple GP models instead of single GP surrogate in CLBO. 
After that, we discuss the related studies to highlight the contribution of our work. 
Also note that, though our work can be combined with various acquisition functions, we focus on BO algorithms that use the expected improvement (EI) as the acquisition function.

\subsection{Basics and Issues of EI-based Bayesian Optimization}
Since GP and acquisition function constitute the key ingredients of BO, we introduce them briefly in this subsection.
In particular, we discuss the reason why using multiple GP models based on the analysis of the ``over-exploitation" and ``over-exploration" issues of EI.

\subsubsection{Standard Gaussian Process}
GP is a stochastic process to approximate the input-output relation, which formulates the function prediction $y({\bf{x}})$ as a Gaussian distribution~\cite{rasmussen2010gaussian} :
\begin{equation}
\begin{scriptsize}
y({\bf{x}}) \sim {\rm{GP}}(m({\bf{x}}),k({\bf{x}},{\bf{x}}'))
\end{scriptsize}
\end{equation}
where, $m({\bf{x}})$ and $k({\bf{x}},{\bf{x}}')$ denote the mean and covariance function, respectively. 
Without loss of generality, we take $m({\bf{x}}) \equiv 0$ and use the popular squared exponential (SE)~\cite{2021Revisiting} covariance function to describe the covariance function:
\begin{equation}
\begin{scriptsize}
{k_{SE}}\left( {{\bf{x,x}}'} \right) = \sigma _f^2\exp \left( { - \frac{1}{2}{{\left( {{\bf{x}} - {\bf{x}}'} \right)}^T}{P^{ - 1}}\left( {{\bf{x}} - {\bf{x}}'} \right)} \right)
\end{scriptsize}
\end{equation}
where, $\sigma _f^2$ models the function output scale, $P$ is an $m \times m$ diagonal matrix with diagonal element  ${\left\{ {l_h^2} \right\}_{1 \le h \le d}}$ , $d$ is the dimension of design space, and $l_h^2$ is the characteristic length that describes how quickly the function value changes as ${\bf{x}}'$ moves away from ${\bf{x}}$ along dimension $h$.

Given the samples ${\bf{y}}(X) = {[y({{\bf{x}}_1}),y({{\bf{x}}_2}) \cdots y({{\bf{x}}_n})]^T}$, the mean prediction and corresponding mean squared prediction error ${\sigma ^2}$ at an unknown site ${{\bf{x}}}$ are expressed as:
\begin{equation}
\begin{small}
\begin{aligned}
\hat y({{\bf{x}}}) &= {\bf{k}}({{\bf{x}}},X){\left( {K + \sigma _n^2{\bf{I}}} \right)^{ - 1}}{\bf{y}}\\
{{\hat \sigma }^2}({{\bf{x}}}) &= {\bf{k}}({{\bf{x}}},{{\bf{x}}}) - {\bf{k}}({{\bf{x}}},X){\left( {K + \sigma _n^2{\bf{I}}} \right)^{ - 1}}{\bf{k}}(X,{{\bf{x}}}) + \sigma _n^2
\end{aligned}
\end{small}
\end{equation}
where, $\sigma _n^2$ is the Gaussian noise, ${\bf{k}}$ is the cross-correlation vector between ${{\bf{x}}}$ and $X$, and $K$ is the covariance matrix over the input features of samples $X$, which are formulated as:
\begin{equation}
\begin{small}
\begin{aligned}
k({\bf{x}},X) &= [k({\bf{x}},{{\bf{x}}_1}), \cdots ,k({\bf{x}},{{\bf{x}}_n})]\\
K &= \left[ {\begin{array}{*{20}{c}}
{k({{\bf{x}}_1},{{\bf{x}}_1})}& \cdots &{k({{\bf{x}}_1},{{\bf{x}}_n})}\\
{}& \ddots & \vdots \\
{Sym}&{}&{k({{\bf{x}}_n},{{\bf{x}}_n})}
\end{array}} \right]
\end{aligned}
\end{small}
\end{equation}
where, \begin{small}$Sym$\end{small} in Eq.(6) means that $K$ is a symmetric matrix. 

Note that the form of ${\bf{k}}$ and $K$ in Eq.(6) can be adjusted in various forms to achieve even better surrogate accuracy in BO process. 
However, no single model can be guaranteed to be most accurate for all the scenarios, therefore it is attractive to employ multiple models instead of a single GP surrogate.
In the meantime, according to Eq.(1), purely increasing the model diversity can increase the error of individual models (i.e., $E_{ind}$), resulting in even worse overall surrogate accuracy.
Therefore, inspired by agreement-assisted co-training algorithms, we propose to exploit unlabeled information and enforce model agreement constraint to improve the overall surrogate accuracy in BO process.


\subsubsection{Over-Exploitation and Over-Exploration of EI}
Assuming that the GP prediction $Y({\bf{x}})$ follows the distribution as \begin{small}$Y({\bf{x}}) \sim N(\hat y({\bf{x}}),{\sigma ^2}({\bf{x}}))$\end{small}, and let \begin{small}${F_{Y({\bf{x}})}}$\end{small} denote the probability distribution of $Y({\bf{x}})$, and \begin{small}${F'_{Y({\bf{x}})}}$\end{small} be the derivative of \begin{small}${F_{Y({\bf{x}})}}$\end{small}, the expectation of the objective improvement can be derived as:
\begin{equation}
\begin{small}
\begin{aligned}
EI({\bf{x}}) &= \int_{ - \infty }^\infty  {\left( {{f_{\min }} - Y({\bf{x}}),0} \right)} d{F_{y({\bf{x}})}}\\
&= \int_{ - \infty }^{{f_{\min }}} {\left( {{f_{\min }} - \hat y} \right)} d{F_{y({\bf{x}})}} - \int_{ - \infty }^{{f_{\min }}} {\left( {y - \hat y} \right)} d{F_{y({\bf{x}})}}\\
&= \left( {{f_{\min }} - \hat y} \right){F_{y({\bf{x}})}}\left( {{f_{\min }}} \right) + \sigma {{F'}_{y({\bf{x}})}}\left( {{f_{\min }}} \right)
\end{aligned}
\end{small}
\end{equation}
In each optimization cycle, we select the sample to query by maximizing EI. The partial derivatives of EI with respect to \begin{small}$\hat y$\end{small} and \begin{small}$\sigma $\end{small} and the gradient of EI are:
\begin{equation}
\begin{small}
\begin{aligned}
\frac{{\partial \left( {EI} \right)}}{{\partial \hat y}} &=  - {F_{Y({\bf{x}})}}\left( {{f_{min}}} \right) =  - \Phi \left( {\frac{{{f_{min}} - \hat y}}{\sigma }} \right) = - \Phi (z)\\
\frac{{\partial \left( {EI} \right)}}{{\partial \sigma }} &= {{F'}_{Y({\bf{x}})}}\left( {{f_{min}}} \right) = \phi \left( {\frac{{{f_{min}} - \hat y}}{\sigma }} \right) =\phi (z)
\end{aligned}
\end{small}
\end{equation}

\begin{equation}
\begin{small}
\begin{aligned}
 - \nabla EI({\bf{x}}) &=  - \left( {\frac{{\partial EI}}{{\partial \hat y}}\frac{{\partial \hat y}}{{\partial {\bf{x}}}} + \frac{{\partial EI}}{{\partial \sigma }}\frac{{\partial \sigma }}{{\partial {\bf{x}}}}} \right)\\
 &= \Phi (z)\frac{{\partial \hat y}}{{\partial {\bf{x}}}} - \phi (z)\frac{{\partial \sigma }}{{\partial {\bf{x}}}}
\end{aligned}
\end{small}
\end{equation}
As shown in Eq.(8), there is a balance in between \begin{small}$\partial \hat y/\partial \bf{x}$\end{small} and \begin{small}$\partial \sigma/\partial \bf{x} $\end{small} to attain maximum EI, which can be varied according to the corresponding weights, i.e., \begin{small}$\Phi (z)$\end{small} and \begin{small}$\phi (z)$\end{small}. 
\begin{figure}[t!]
	\centering
	\includegraphics[width=3.5in]{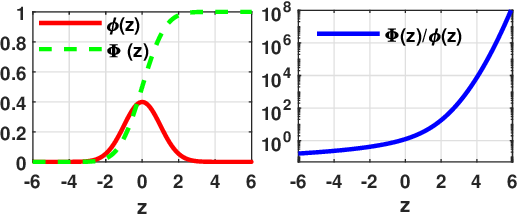}
	\caption{Plots of $\Phi (z)$ and $\phi (z)$ and ${{\Phi (z)} \mathord{\left/
				{\vphantom {{\Phi (z)} {\phi (z)}}} \right.
				\kern-\nulldelimiterspace} {\phi (z)}}$.}
	\label{fig:figure2-ai}
\end{figure}

In particular, the influences of \begin{small}$\Phi (z)$\end{small} and \begin{small}$\phi (z)$\end{small} on the performance of EI are discussed with Fig.1.
When \begin{small}$z>3$\end{small}, \begin{small}$\Phi (z)$\end{small} can be overwhelmingly greater than that of \begin{small}$\phi (z)$\end{small}.
Correspondingly, the following equation can be derived:
\begin{equation}
\begin{small}
\begin{aligned}
 - \nabla EI({\bf{x}}|z > 3) &= \Phi (z > 3)\frac{{\partial \hat y}}{{\partial {\bf{x}}}} - \phi (z > 3)\frac{{\partial \sigma }}{{\partial {\bf{x}}}}\\
 &\approx \Phi (z > 3)\frac{{\partial \hat y}}{{\partial {\bf{x}}}}\\
\end{aligned}
\end{small}
\end{equation}
It means, EI focuses sampling near the optima of the fitted GP surrogate when $z$ takes large positive values, and thereby it may ignore exploring other regions of the optimization space and thus get stuck in a sub-optimal solution, which is the well-known issue of EI labeled as ``over-exploitation''.

On the other hand, as shown in Fig.1, \begin{small}$\phi (z)$\end{small} gradually becomes greater than \begin{small}$\Phi (z)$\end{small} when $z<0$.
In particular, we can ``approximately'' have the following equation when \begin{small}$z<-3$\end{small}:
\begin{equation}
\begin{small}
\begin{aligned}
 - \nabla EI({\bf{x}}|z <  - 3) &= \Phi (z <  - 3)\frac{{\partial \hat y}}{{\partial {\bf{x}}}} - \phi (z <  - 3)\frac{{\partial \sigma }}{{\partial {\bf{x}}}}\\
 &\approx -\phi (z <  - 3)\frac{{\partial \sigma }}{{\partial {\bf{x}}}}\\
\end{aligned}
\end{small}
\end{equation}
This happens at later stage of the optimization process when no objective improvement can be expected directly from the GP surrogate. At such stage with $z<0$, the issue of "over-exploration" can take place, i.e., EI is likely to query samples with maximum \begin{small}$\sigma ({\bf{x}})$\end{small} but poor \begin{small}$y({\bf{x}})$\end{small} for a long period of time.

More specifically, the ``over-exploration" issue can be explained from a historical view of the optimization process, with the following equation:
\begin{equation}
\begin{small}
\begin{aligned}
&z = \frac{{{f_{min}} - \hat y({\bf{x}})}}{{\sigma ({\bf{x}})}} > 0 \Rightarrow \hat y({\bf{x}}_{EI}^*) < {f_{min}}\\
&z = \frac{{{f_{min}} - \hat y({\bf{x}})}}{{\sigma ({\bf{x}})}} \le 0 \Rightarrow No\;constraints\;on\;\hat y({\bf{x}}_{EI}^*)\;
\end{aligned}
\end{small}
\end{equation}
where, ${\bf{x}}_{EI}^*$ denotes the selected sample to query by maximizing EI, and \begin{small}$\hat y({\bf{x}}_{EI}^*)$\end{small} is the related GP prediction.
Note that the EI-based BO process often starts with $z>0$. 
As shown in Eq.(11), the condition \begin{small}$z>0$\end{small} confines \begin{small}$\hat y({\bf{x}}_{EI}^*)$\end{small} to be queried in areas with relatively better \begin{small}$
\hat y({\bf{x}})$\end{small}.
In the meantime, the GP surrogate can more or less capture the global trend of objective function.
Therefore, as the iteration goes on with $z>0$, the density of samples at regions with relatively better \begin{small}$y({\bf{x}})$\end{small} becomes denser than those with poor \begin{small}$y({\bf{x}})$\end{small}.
After that when $z\le0$, EI turns to exploring the optimization space globally by putting much greater weight on \begin{small}$\sigma(\bf{x})$\end{small} (see Eq.(10).
However, the condition \begin{small}$z \le 0$\end{small} does not impose any constraints on \begin{small}$\hat y({\bf{x}}_{EI}^*)$\end{small} (see Eq.(11)).
Then, \emph{the ``over-exploration" issue takes place, and EI favors querying samples with 
maximum \begin{small}$\sigma ({\bf{x}})$\end{small} but poor \begin{small}$y({\bf{x}})$\end{small} continuously and even forever}~\cite{2021Guo}.

To summarize, the parameter $z$, which is a function of $\hat y(\bf{x})$ and $\sigma(\bf{x})$, plays a crucial role that determines the behavior of EI and thus the performance of BO. 
While increasing $\sigma(\bf{x})$ can help to address the issue of ``over-exploitation" (i.e., the issue with $z>0$), larger $\sigma(\bf{x})$ can also lead the algorithm get stuck in the ``over-exploration" issue more easily, which takes place when $z<0$.

Intrinsically, the ``over-exploitation" and ``over-exploration" can attribute to the poor accuracy of the trained GP surrogate, particularly that in the region where the true optima locate.
Therefore, we propose to use multiple GP models to improve the overall surrogate accuracy in CLBO.
And further, we propose to make use of both the model diversity and agreement on unlabeled information to manage the overall accuracy of multiple GP models in BO process, which are illustrated with more details in Section III.

\subsection{Related Work}

Our proposed CLBO can have relations with the following categories of BO algorithms.
First, CLBO is most close to BO algorithms assisted by multiple GP kernels~\cite{martinez2018funneled} or a set of surrogate techniques~\cite{2013Efficient}, as they all attempt to take advantage of the model differences to complement each other and thus achieving even better optimization results.
Differently, in addition to model differences, CLBO also makes use of the model consistency under the spirit of co-training to manage the overall prediction accuracy in BO process.
Second, CLBO shares similarity with batch BO algorithms, such as constant liar~\cite{ginsbourger2010kriging}, batch BO via local penalization~\cite{J2015Batch}, etc., since they both query more than one samples in each optimization cycle.
More importantly, the querying samples in each cycle of the above studies are obtained by maximizing a modified EI function.
In other words, instead of using multiple GP surrogates, these studies can be regarded as BO algorithms that assisted by multiple acquisition functions. 
Third, CLBO can also have relation to the BO algorithms that augment the role of ${\sigma}\left( {\bf{x}} \right)$~\cite{1998Global},~\cite{ponweiser2008clustered} or change the incumbent~\cite{berk2018exploration} of EI formulation, as they share the same purpose to resolve the ``over-exploitation" issue of EI.
However, some of the above studies may not be helpful to address the ``over-exploration" issue of EI~\cite{2021Guo}.
Differently, by exploiting both model agreement and disagreement (i.e., diversity) to improve the overall surrogate accuracy, our work can help to resolve the issues of ``over-exploitation" and ``over-exploration" simultaneously.


\section{Methodology}

To address the suboptimal problem of BO and therefore achieving more efficient global optimization, we propose CLBO which exploits both model diversity and agreement on unlabeled information to improve the overall surrogate accuracy in the optimization process.
In this section, we show the details of proposed CLBO.

\subsection{General Framework of CLBO}

The basic idea of CLBO is as follows: Firstly, while no surrogate can be guaranteed to be most accurate at all sites of the design space, multiple surrogate models can complement each to achieve better approximation accuracy particular that in neighborhood of the true optimal solution, therefore we propose to build multiple GP predictions with the subsets of training samples in BO process.
Secondly and more importantly, the GP models trained with a small portion of subset samples may suffer from poor individual accuracy and therefore resulting in even worse overall surrogate accuracy (see Eq.(1)). Nonetheless, enforcing model agreement on unlabeled information can reduce the number of samples necessary to achieve required model accuracy~\cite{leskes2008value}. Hence, we propose to exploit model agreement to ensure reasonable accuracy of individual models. 
Then, these relatively accurate individual models are combined to complement each other and thus achieving better overall approximation accuracy in BO process.
Along this idea, we show the framework of proposed CLBO in Algorithm 1, which is consisted of four components, i.e., (1) generation of diverse sample subsets; (2) building of agreement-assisted GP models; (3) sample search with multiple GP models and (4) sample exchange with multiple GP models.

Note that the building of agreement-assisted GP models is the most important part of CLBO, and there can be many ways to exploit model agreement on unlabeled knowledge.
In the following subsections, we present an instantiation of agreement-assisted GP models. 
After that, we illustrate the details of the other components of CLBO.

\begin{algorithm}[t]
	\caption{General framework of CLBO}
	\begin{algorithmic}[1]
		\REQUIRE{Target optimization problem and the number of initial training samples}
		
		\ENSURE{Optimal solution of the target optimization problem}
		\STATE \textbf{Generation of Diverse Sample Subsets:} Use bootstrapping (see Algorithm 2) or other sampling strategy to obtain a set of sample sets, and evaluate the objective function value.\\

		\WHILE{the termination condition is not satisfied}
		\STATE \textbf{Building of Agreement-Assisted GP Models:}
		Build multiple GP models, where agreement constraints are imposed on all or part of GP models in the training process to obtain predictions of reasonably accuracy.
		\STATE \textbf{Sample Search with Multiple GP Models:} Use acquisition functions such as EI to select the samples to query, and then evaluate the objective functions of all the new samples.
		\STATE \textbf{Sample Exchange between Multiple GP Models:} Add new samples to each training set of the multiple models.
		
		\ENDWHILE

	\end{algorithmic}
\end{algorithm}

		

		


\subsection{Agreement-Assisted GP Models}

According to the bias-variance tradeoff equation (see Eq.(1)), it is desirable to combine individual models of reasonable accuracy to complement each, and therefore achieving better overall surrogate accuracy in BO process.
To achieve it, we use two kinds of GP models to deal with the balance in between individual model accuracy and diversity among models.
One is the single-output GP (SOGP) that trained with full sample set, which is believed to have relatively good mean accuracy over design space. 
The other kind of GP models are built by using multi-output GP (MOGP)~\cite{liu2018remarks},~\cite{durichen2014multi} with training subsets, wherein agreement constraint are imposed on curve bumpiness to ensure reasonable model accuracy.
As we can train the SOGP by following the standard process shown in Section II.A, this section will be focused on illustrating the motivation and implementing details of agreement assisted MOGP.

Specifically, the agreement assisted MOGP built in this paper is inspired by the multi-task GP (MTGP)~\cite{bonilla2008multi}, which is originally proposed for transfer learning between distinct tasks~\cite{pan2009survey},\cite{luo2018evolutionary}.
In this paper, by treating the subsets as related source data, we extend the MTGP to generate multiple GP models with subsets.
The motivation behind is as follows. That is, as the surrogate models trained with subsets are built for the same task, they should have similar curvature/curve bumpiness in predicting the underlying objective function in unexplored design space.
To distinguish it from the multi-task case, we name it as multi-form GP (abbreviated as MFGP).

In MFGP, let Eq. (13) denote the training subsets to build MFGP, 
we use a correlation matrix to model the relations between them, where ${\rho _{ij}}$ in Eq.(14) denotes the correlation between the outputs associated with ${X_i}$ and ${X_j}$.

\begin{equation}
\begin{small}
\begin{aligned}
X &= \left[ {{X_1}, \cdots ,{X_j} \cdots {X_m}} \right],{X_1} \ne  \cdots  \ne {X_j} \ne  \cdots  \ne {X_m}\\
{\bf{y}} &= {\left[ {{\bf{y}}_1^{}, \cdots ,{\bf{y}}_j^{}, \cdots {\bf{y}}_m^{}} \right]^T},\;{\bf{y}}_1^{} \ne  \cdots  \ne {\bf{y}}_j^{} \ne  \cdots  \ne {\bf{y}}_m^{}
\end{aligned}
\end{small}
\end{equation}
\begin{equation}
\begin{scriptsize}
{K^f} = \left[ {\begin{array}{*{20}{c}}
	{{\rho _{11}}}& \cdots &{{\rho _{1m}}}\\
	\vdots & \ddots & \vdots \\
	{{\rho _{m1}}}& \cdots &{{\rho _{mm}}}
	\end{array}} \right]\;\;where\;{\rho _{ij}} = {\rho _{ji}}
\end{scriptsize}
\end{equation}
Then, we reformulate the cross-correlation vector ${k_j}$ and the covariance matrix ${K_M}$ as:
\begin{equation}
\begin{small}
\begin{aligned}
{K_M} = \left[ {\begin{array}{*{20}{c}}
{{\rho _{11}}{K_{11}}}& \cdots &{{\rho _{1m}}{K_{1m}}}\\
 \vdots & \ddots & \vdots \\
{{\rho _{m1}}{K_{m1}}}& \cdots &{{\rho _{mm}}{K_{mm}}}
\end{array}} \right] + D \otimes I
\end{aligned}
\end{small}
\end{equation}
where, ${\{ {K_{ij}}\} _{1 \le i,j \le m}}$ is the input correlation matrix associated with $X_i$ and $X_j$, and $D$ is an $m \times m$ diagonal matrix with diagonal elements \begin{small}${\left\{ {\sigma _{n,j}^2} \right\}_{1 \le j \le m}}$\end{small}, $\sigma _{n,j}^2$ is the noise term of the GP associated with ${X_j}$. Then, the GP predictions at ${{\bf{x}}}$ can be expressed as: 
\begin{equation}
\begin{small}
\begin{aligned}
{\bf{\hat y}}({{\bf{x}}}) &= \left[ {\begin{array}{*{20}{c}}
	{{k_1}({{\bf{x}}},X)K_M^{ - 1}{\bf{y}}}\\
	\vdots \\
	{{k_m}({{\bf{x}}},X)K_M^{ - 1}{\bf{y}}}
	\end{array}} \right]\\
{{\bf{\sigma }}^{{2}}}({{\bf{x}}}) &= \left[\! {\begin{array}{*{20}{c}}
	{{k_1}({{\bf{x}}},{{\bf{x}}}) - {k_1}({{\bf{x}}},X)K_M^{ - 1}{k_1}(X,{{\bf{x}}}) \!+\! \sigma _{n,1}^2}\\
	\vdots \\
	{{k_m}({{\bf{x}}},{{\bf{x}}})\! - \!{k_m}({{\bf{x}}},X)K_M^{ - 1}{k_m}(X,{{\bf{x}}}) + \sigma _{n,m}^2}
	\end{array}} \!\right]
\end{aligned}
\end{small}
\end{equation}

Also note that, in original MTGP~\cite{bonilla2008multi}, the correlation coefficient ${\rho _{ij}}$ also measures the differences between source tasks to avoid “negative transfer"~\cite{wei2017source},~\cite{da2018curbing}. 
In our MFGP, we use ${\rho _{ij}}$ to model the differences of function outputs associated with subsets, which plays a crucial role in generating different models. 
To ensure the diversity of GP models,
we generate subsets by using algorithms in Subsection A, and we also take ${\rho _{ij}}$ and $\sigma _{n,i}^2$ as hyper-parameters in the tuning process.

Equation (17) shows the hyper-parameters of MFGP to be tuned by the negative logarithm maximum likelihood (NLML)~\cite{liu2018remarks}. 
As contrast,  Eqs. (18) and (19) show the GP models of subsets trained by SOGP, where $\sigma _{f,i}^2$ models the function output scale associated with subset\begin{small} ${(X_i,{\bf{y}}_i)}$\end{small}.
\begin{equation}
\begin{small}
\begin{aligned}
\arg \min \;NLM{L_{MFGP}}\left( {l_h^2,\sigma _{f,i}^2,\sigma _{n,i}^2,{\rho _{i,j}}} \right)\\
\end{aligned}
\end{small}
\end{equation}
\begin{equation}
\begin{small}
\begin{aligned}
\hat y_i({{\bf{x}}_q}) &= {\bf{k}}({{\bf{x}}_q},X_i){\left( {K(X_i) + \sigma _{n,i}^2{\bf{I}}} \right)^{ - 1}}{\bf{y_i}}\\
{{\hat \sigma_i }^2}({{\bf{x}}_q}) &= {\bf{k}}({{\bf{x}}_q},{{\bf{x}}_q}) \\
&- {\bf{k}}({{\bf{x}}_q},X_i){\left( {K(X_i) + \sigma _{n,i}^2{\bf{I}}} \right)^{ - 1}}{\bf{k}}(X_i,{{\bf{x}}_q}) + \sigma _{n,i}^2
\end{aligned}
\end{small}
\end{equation}

\begin{equation}
\begin{scriptsize}
\arg \min \;NLM{L_{SOGP}}(l_{h,i}^2,\sigma _{f,i}^2,\sigma _{n,i}^2)
\end{scriptsize}
\end{equation}

More specifically, $l_h^2$ in Eq.(17) models the curve bumpiness of all the subsets in MFGP. 
In contrast, in Eq. (19), $l_{h,i}^2$ is the length hyper-parameter of each subset.
In other words, compared to SOGP shown in Eq. (19), an agreement constraint on curve bumpiness has been imposed among the GP models in MFGP, which can be written as below:.  
\begin{equation}
\begin{small}
\underbrace {l_{h,1}^2 =  \cdots l_{h,i}^2 =  \cdots  = l_{h,m}^2 \equiv l_h^2}_{Agreement\;constraint\;on\;curve\;bumpiness\;in\;MFGP}
\end{small}
\end{equation}
The effectiveness of the agreement constraint in MFGP will be verified in Section IV.

\begin{algorithm}[t]
	\caption{Bootstrap sampling}
	\begin{algorithmic}[1]
		\REQUIRE{The full training set $X$, the number $n$ of samples in $X$ and the number $m$ of subsets}
		
		\ENSURE{The subsets ${X_1}, \cdots ,{X_m}$ }
		\FOR{$i \le m$}
		\STATE Bootstrapping $n$ samples from $X$ to obtain ${X_{sub}}$ 
		\STATE Use the unique function to delete the duplicate samples in ${X_{sub}}$ to get ${X_i}$, i.e., ${X_i} = unique\left( {{X_{sub}}} \right)$
		\ENDFOR
		
	\end{algorithmic}
\end{algorithm}

\subsection{Sample Search with Multiple GP Models} 


After building multiple models using MFGP and SOGP, we use EI to search new samples. Besides, we set a minimum distance threshold (i.e., ${d_{\min }}<\varepsilon$) to avoid the ill-conditioning of the inversion of correlation matrix. To prevent samples being null due to the threshold in an optimization cycle, we use PEI~\cite{zhan2017pseudo} as a remedy for the search with subsets. The PEI can be expressed as:
\begin{equation}
\begin{small}
\begin{aligned}
PEI({\bf{x}},{{\bf{x}}^*}) &= EI({\bf{x}}) \times IF({\bf{x}},{{\bf{x}}^*})\\
IF({\bf{x}},{{\bf{x}}^{\rm{*}}}) &= 1 - \exp \left( { - \sum\limits_{h = 1}^d {{{{{\left\| {x_h^{} - x_h^*} \right\|}^{\rm{2}}}} \mathord{\left/
				{\vphantom {{{{\left\| {x_h^{} - x_h^*} \right\|}^{\rm{2}}}} {2{l_h}^2}}} \right.
				\kern-\nulldelimiterspace} {2{l_h}^2}}} } \right)
\end{aligned}
\end{small}
\end{equation}
where, ${{\bf{x}}^*}$ is the initially selected sample of $maxEI$ and $l_h^2$ is the characteristic length of covariance function in Eq.(4). 
According to our observations, PEI will only be called in rare cases. More details are shown in Algorithm 3.

\begin{algorithm}[t]
	\caption{Sample search with multiple GP models}
	\begin{algorithmic}[1]
		\REQUIRE{The trained SOGP and MFGP models, the number $m$ of subsets}
		\ENSURE{The new query samples, i.e.,\begin{small} $X_{new}$ \end{small}}
		\STATE Optimize the EI acquisition function with the SOGP model, i.e., \begin{small}${X_{new}^{SOGP}} = \arg \max EI({\bf{x}})$\end{small}
		\FOR{\begin{small} $i \le m$\end{small}}
		\STATE Optimize the EI with the ${i^{th}}$ sub-surrogate in MFGP, i.e., \begin{small}${{\bf{x}}_{new}} = \arg \max EI({\bf{x}})$\end{small}
		\STATE  Check the minimum distance condition for  ${\bf{x}}_{new}^{}$, i.e.,\\\begin{small}${d_{\min }} = \min \left\| {{{\bf{x}}_{new}} - X \cup X_{new}^{SOGP} \cup X_{new}^{MOGP}} \right\|$\end{small}
		\IF{\begin{small} ${d_{\min }} < \varepsilon $ \end{small}}
		\STATE \begin{small} ${{\bf{x'}}_{new}} = \arg \max PEI({\bf{x}},{{\bf{x}}_{new}})$ \end{small};\\ \begin{small}${{\bf{x}}_{new}} = {{\bf{x'}}_{new}}$\end{small}
		\ENDIF
		\STATE $X_{new}^{MFGP} = X_{new}^{MFGP} \cup {\bf{x}}_{new}^{}$
		\ENDFOR
		\STATE $X_{new} = X_{new}^{MFGP} \cup {\bf{x}}_{new}^{SOGP}$
		
	\end{algorithmic}
\end{algorithm}

\subsection{Sample Exchange between Multiple GP models} 
On one hand, sample exchanges will provide clues to each GP model about where the real optimum may locate; 
on the other hand, we should keep model diversity to gain maximum improvement by the co-training style process. 
Therefore, we randomly assign the new sample of SOGP model to a MFGP subset if it was inferior to the current best solution ${f_{\min }}$, otherwise it will be added to each MFGP subset. In the meantime, we directly add the new querying samples of MFGP models to the SOGP training set. The implementation details are shown in Algorithm 4.
\begin{algorithm}[t]
	\caption{Sample exchange and update of training set}
	\begin{algorithmic}[1]
		\REQUIRE{The new sample points as \begin{small} ${X_{new}} = {\bf{x}}_{new}^{SOGP} \cup X_{new}^{MFGP}$ \end{small}}
		
		\ENSURE{The updated training sets of SOGP and MFGP}
		
		\STATE Evaluate ${X_{new}}$ in parallel with the real objective function
		\STATE Update the current best optimal solution, \\ 
		$i.$ \begin{small} ${f_{\min }} \leftarrow \min \left( {{f_{\min }},Y\left( {{X_{new}}} \right)} \right)$ \end{small} \\
		$ii.$ \begin{small}${{\bf{x}}_{\min }} \leftarrow {\bf{x}} \in X:y({\bf{x}}) = {f_{\min }}$ \end{small}
		
		\STATE Update the training set of SOGP \\
		$i.$ \begin{small} $X \leftarrow X \cup {X_{new}}$  \end{small}\\
		$ii.$\begin{small} $Y \leftarrow Y \cup {Y_{new}}$ \end{small}\\
		
		\STATE Update the training set of MFGP
		
		\IF{${{\bf{x}}_{\min }}$ comes from ${\bf{x}}_{new}^{SOGP}$}
		\FOR{ \begin{small} $i \le m$ \end{small}}
		\STATE \begin{small} ${X_i} \leftarrow {X_i} \cup {\bf{x}}_{new}^{SOGP}$ \end{small}
		\STATE \begin{small} ${Y_i} \leftarrow {Y_i} \cup Y_{new}^{SOGP}$ \end{small}
		\ENDFOR
		\ELSIF{${{\bf{x}}_{\min }}$ comes from ${{X}}_{new}^{MOGP}$ }
		\STATE \begin{small} $nbest = \arg \min (Y_{new}^{MOGP})$ \end{small}
		\FOR {$i \le m\;\& \;i \ne nbest$}
		\STATE \begin{small} ${X_i} \leftarrow {X_i} \cup X_{new}^{MFGP}(nbest)$ \end{small}
		\STATE \begin{small} ${Y_i} \leftarrow {Y_i} \cup Y_{new}^{MFGP}(nbest)$ \end{small}
		\ENDFOR
		\STATE Randomly choose ${X_j}$
		\STATE \begin{small} ${X_j} \leftarrow {X_j} \cup {\bf{x}}_{new}^{SOGP}$ \end{small}
		\STATE \begin{small} ${Y_j} \leftarrow {Y_j} \cup Y_{new}^{SOGP}$ \end{small}
		\ENDIF
		
	\end{algorithmic}
\end{algorithm}

\section{Experiments}

In this section, we test CLBO and compare it with both classical and recent BO algorithms on numerical and engineering benchmark problems.
After that, we examine the influence of different GP models on CLBO performance.

\subsection{Baselines for comparison}

To show the effectiveness of CLBO, it is compared against the classical EGO~\cite{jones1998efficient} and the following three kinds of BO algorithms:

(1) To verify the effectiveness of exploiting both model diversity and agreement on unlabeled information in surrogate management for BO, CLBO is compared with algorithms that purely take advantage of the differences between models to improve algorithm performance. 
In particular, an efficient global optimization algorithm assisted by multiple surrogate techniques (labeled as MSEGO~\cite{2013Efficient}) is taken as a baseline.
In MSEGO, the predictions from surrogates such as radial basis function (RBF), support vector machine (SVM), etc. are combined with the uncertainty imported from GP to formulate new EI acquisition functions, and then a set of samples are queried  in each optimization cycle by maximizing the new EI functions.
Additionally, we also build a algorithm labeled as MSBO, which trains the surrogates with the full and subset samples using standard GP while remaining the other settings exactly the same as those of CLBO.

(2) To show the effectiveness of improving GP surrogate accuracy other than modifying the acquisition function to address the suboptimal problem of BO, CLBO is compared against two variants of EI-based algorithms, i.e., generalized EI (GEI)~\cite{ponweiser2008clustered} and Exploration Enhanced EI (E3I)~\cite{berk2018exploration}.
Specifically, GEI proposes to address the ``over-exploitation" issue of EI (see Eq.(10)) by dynamically adjusting the weight of $\sigma(\bf{x})$ with a global factor $g$;
and E3I proposes to use the optima of Thompson samples to replace the current best solution \begin{small}$f_{min}$\end{small} as the incumbent in EI, and therefore lowering the weight of \begin{small}$\partial \hat y/\partial \bf{x}$\end{small} to resolve the ``over-exploitation" when $z>0$.
Additionally, the standard BO with MCMC hyperparameters~\cite{snoek2012practical} is selected as a baseline, labeled as BO-MCMC in the following tests.

(3) Since CLBO evaluates multiple samples in each optimization cycle, it is also compared against a set of batch BO algorithms, such as constant liar (CL)~\cite{ginsbourger2010kriging}, batch BO via local penalization (BBO-LP)~\cite{J2015Batch} and pseudo EI (PEI)~\cite{zhan2017pseudo}.
The common feature of these batch algorithms is that they all rely on a single GP surrogate to guide the optimization process.

\subsection{Implementing Details}

Except BO-MCMC, BBO-LP and MSEGO, CLBO and the other compared baselines are all built based on the GPML toolbox~\cite{rasmussen2010gaussian}.
In particular, the MFGP in CLBO is programmed based on the MTGP toolbox\footnote{https://www.rob.uni-luebeck.de/index.php?id=410\&L=0}, which is an extension of the GPML toolbox.
Additionally, the GpyOpt~\cite{gpyopt2016} is used to run BO-MCMC and BBO-LP.
The source code of MSEGO, which is built upon the SURROGATES toolbox\footnote{https://sites.google.com/site/srgtstoolbox/}, is also employed for the tests.



The number of initial training samples is set as six times of the problem dimension. For numerical benchmark functions, the sample budget is set as 30 times of the problem dimension.
However, the sample evaluation for engineering benchmarks can be quite expensive, and therefore the number of total evaluations for engineering benchmarks is set as no greater than 200. 
Furthermore, since the best solution of numerical benchmark functions are known, the regret~\cite{wang2017max} is used to measure the best solution obtained thus far, i.e., \begin{small}${R_t} = f_{\min }^{(t)} - {\min _{{\bf{x}} \in \chi }}f({\bf{x}})$\end{small}, wherein \begin{small}${\min _{{\bf{x}} \in \chi }}f({\bf{x}})$\end{small} is the real optimal solution of the benchmarks, and $f_{\min }^{(t)}$ denotes the best solution at the ${t^{th}}$ iteration.
For engineering benchmarks, the best objective function values obtained thus are directly used to show the convergence history of compared algorithms.

By default, we use MFGP to obtain two GP predictions with training subsets, and one single-output GP with the full samples in CLBO. The influence of different GP models on CLBO performance are further inspected in Subsection E. 
Accordingly, three samples are evaluated in each optimization cycle when running batch BO algorithms such as CL, PEI and BBO-LP.
By following the demo shown in SURROGATE toolbox, the radial basis function (RBF) and linear Shepard function are used to combine with standard GP for MSEGO.

Although the algorithm efficiency can be boosted by evaluating a set of samples simultaneously, the main purpose of this paper is to address the suboptimal problem of BO. 
Therefore, the performance of sequential and batch BO algorithms are compared by the convergence history in  terms of function calls within a fixed sample budget. 
And accordingly, the termination conditions for the compared algorithms are set as follows:

(1) $n_{total} \le n_{T}$, i.e., the total number of function calls $n_{total}$ cannot be greater than the sample budget $n_{T}$.

(2) $t_{total} \le t_{T}$, i.e., the total number of iterations $t_{total}$ should be no greater than a threshold $t_{T}$.

In particular, the condition $t_{total} \le t_{T}$ is set to address the following issue.
That is, to avoid the ill-conditioning of the correlation matrix a distance constraint \begin{small}$min\| {{\bf{x}} - {{\bf{x}}_{i}}}\|>\varepsilon$\end{small} is imposed, where ${\bf{x}}_{i}$ is any one of the training samples, and $\varepsilon$ is set to be $0.001$ in the following tests.
Nevertheless, the constraint \begin{small}$min\| {{\bf{x}} - {{\bf{x}}_{i}}}\|>\varepsilon$\end{small} can prevent an algorithm from querying more samples at later stage of the iteration process, and consequently the algorithm may get trapped in infinite loop but cannot meet the termination condition $n_{total} \le n_{T}$.
Therefore, we propose to terminate the algorithm when either the condition (1) or (2) is met.

\subsection{Comparisons on Numerical Tests} 
Five numerical benchmark functions are selected from~\cite{simulationlib} to inspect the algorithm performance.
Specifically, Hartman6 is a non-convex problem that frequently used as a benchmark for testing BO algorithms~\cite{martinez2018funneled},~\cite{2021Guo}; Michalewicz is a multimodal function characterized by steep ridges; and Rastrigin and Ackley are functions featured  by many local optima.  
The Michalewicz, Rastrigin and Ackley are tested over five-dimensional design spaces.
The steepness parameter $m$ of Michalewicz is set to be 10.
While the variable ranges of Michalewicz and Rastrigin are set according to the descriptions in~\cite{simulationlib}; the range of each variable in Ackley is set as [-2, 2], by referring to~\cite{zhang2018variable}. 
Additionally, the 10-dimensional Trid function is also selected for the test. 
As the function values of Trid vary in several orders of magnitude, it cannot be easily fitted with a small portion of samples.


Figures 2 to 6 show the testing results on numerical benchmarks,
where the shaded areas in Figs.2(a) to 6(a) and Figs. 2(b) to 6(b) show the results variation of different algorithms over 20 runs.
The boxplots in Figs.2(c) to 6(c) exhibit the medians and the distributions of the final optimal solutions.
Obviously, CLBO always achieves the best or the second best optimal solutions within sample budget. 

\begin{figure}[t!]
	\centering
	\includegraphics[width=3in]{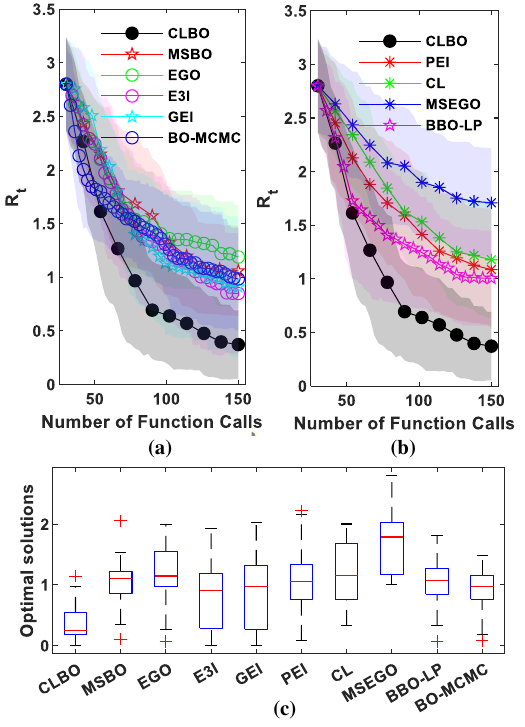}
	\caption{Test results of five-dimensional Michalewicz function, (a) comparison with sequential BO algorithms, (b) comparison with batch BO algorithms,
	(c) boxplot of the regrets of final optimal solutions}
	\label{fig:figure2-ai}
\end{figure}

\begin{figure}[t!]
	\centering
	\includegraphics[width=3in]{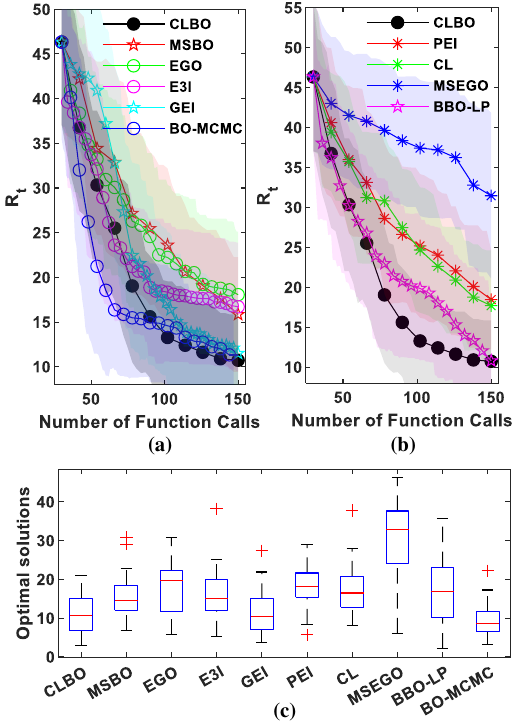}
	\caption{Test results of five-dimensional Rastrigin function, (a) comparison with sequential BO algorithms, (b) comparison with batch BO algorithms,
	(c) boxplot of the regrets of final optimal solutions}
	\label{fig:figure2-ai}
\end{figure}

\begin{figure}[t!]
	\centering
	\includegraphics[width=3in]{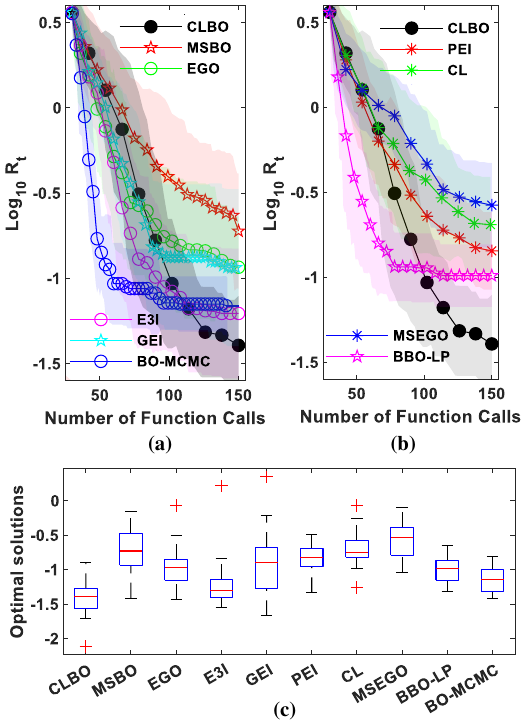}
	\caption{Test results of five-dimensional Ackley function, (a) comparison with sequential BO algorithms, (b) comparison with batch BO algorithms,
	(c) boxplot of the regrets of final optimal solutions}
	\label{fig:figure2-ai}
\end{figure}

When comparing BO-MCMC with CLBO and the other compared algorithms, it shows faster convergence rate at the early stage of the optimization process, especially when optimizing the Rastrigin and Ackley functions.
However, at later stage of the optimization process, CLBO can have better results than those of BO-MCMC.
Among the compared algorithms, E3I performs the best when optimizing the functions such as Trid and Hartman6 functions. 
However, it performs much poorer when testing the Michalewicz and Rastrigin functions. 
The reason behind can be explained as follows.
To address the suboptimal problem of BO, E3I proposes to use the optima of Thompson samples to replace the current best solution \begin{small}$f_{min}$\end{small} as the incumbent, and thereby the value of $z$ becomes smaller in the EI formulation.
In particular, the smaller $z>0$ helps to lower the weight of \begin{small}$\partial \hat y/\partial \bf{x}$\end{small} (see Eq.(10)) and thus addressing the ``over-exploitation'' issue.
However, the smaller $z<0$ will increase the weight of \begin{small}$\partial \sigma/\partial \bf{x} $\end{small} (see Eq.(11)), which can make the algorithm get stuck in the ``over-exploration'' more easily.
For the convex functions like Trid, the optimization stage with $z<0$ can be delayed or even avoided when optimizing it with E3I. 
That is, by reducing the value of $z>0$ to prevent too greedily sampling around the optima of the fitted surrogate, E3I is able to maintain the effective exploitation-exploration tradeoff with $z>0$ and thus achieving better results than the other compared algorithms. 
However, $z<0$ can take place many times for the functions with many local optima, but E3I may even increase the risk of ``over-exploration" when $z<0$. Hence, E3I performs much poorer when optimizing the Michalewicz and Rastrigin functions.

Similar situations can be observed for GEI, which proposes to resolve the ``over-exploitation'' issue by dynamically augmenting the role of \begin{small}$ \sigma (\bf{x}) $\end{small} with a factor $g$.
However, adjusting \begin{small}$ \sigma (\bf{x}) $\end{small} also cannot address the ``over-exploration'' issue with $z<0$.
Moreover, the value of $g$ are set empirically in~\cite{ponweiser2008clustered}, which may not adapt well to the real objective landscape. 
Hence, except Rastrigin function, GEI performs relatively poor when optimizing the benchmark functions.

With the above, it indicates that modifying the EI acquisition function may not be able to fully resolve the suboptimal problem of BO. 
Intrinsically, such suboptimal problem can attribute to the poor accuracy of the trained GP surrogate particularly that in the neighborhood of the real optimal solution.
However, no surrogate can be most accurate at every site of the design space.
Therefore, an alternative way to address the suboptimal problem is to improve the overall surrogate accuracy by using multiple models, which can be promising to address both the issues of ``over-exploitation'' and ``over-exploration''.

However, the performance of MSEGO and MSBO, which make use of multiple surrogate models, are rather poor when optimizing the numerical benchmarks.
According to Eq.(1), the differences between surrogate models indeed can help them to  complement each other, but the surrogates may suffer from incredibly large individual prediction errors, resulting in worse overall prediction accuracy.
In MSEGO and MSBO, no strategy is used to take care of the individual prediction accuracy.
Moreover, the information exchange between the surrogates is limited to exchanging the new queried samples in each optimization cycle. 
Hence, the overall surrogate accuracy particularly that in the vicinity of the real optimal solutions may improve slowly as the iteration goes on, leading to the poor optimization results of MSEGO and MSBO.

Differently, good care of the individual model accuracy and model diversity are taken in CLBO.
\emph{In particular, agreement constraint is imposed on curve bumpiness when training GP predictions with subset samples using MFGP.
As has been proved in~\cite{leskes2008value}, the agreement constraint on unlabeled information helps to reduce the sample complexity of the effective hypothesis space. It means, with the same number of subset samples, the agreement constraint GP predictions can have smaller real prediction loss according to Eq.(2). 
Additionally, one more single-output GP is built with the full sample set, which helps to increase the model diversity with relatively good individual accuracy. 
And  further, the combination of these reasonably accurate GP models can complement each other to achieve even better overall surrogate accuracy (see Eq.(1)) in the vicinity of the true optimal.}
Hence, our proposed CLBO always achieves the best or second best solutions with even faster convergence rate.

As the main purpose of batch BO algorithms such as BBO-LP is to enable parallel optimization without scarifying the global searching performance dramatically, the convergence performance of BBO-LP, PEI and CL look poorer than the sequential BO algorithms in most cases, when the function calls is concerned. 
By using multiple GP models and taking a balance in between individual model accuracy and model diversity, CLBO is the only batch BO algorithm which achieves better convergence rate than the sequential BO algorithms.

To examine the computational costs of the compared algorithms, Table 1 lists the averaged CPU time over 20 runs.
Since the dimensionality of the covariance matrix of MFGP (see Eq.(15)) in CLBO is larger than that of standard GP (see Eq.(6)), the averaged CPU time of CLBO is larger than those of EGO, GEI and etc.
In the meantime, due to the huge cost of Thompson sampling and MCMC, the computational cost of E3I and BO-MCMC can be one order higher than the other compared algorithms. 
As the BO algorithms are often applied to expensive problems, where a function call can take even much longer time to implement, the cost differences of BO algorithms may not be an issue.

\begin{figure}[t!]
	\centering
	\includegraphics[width=3in]{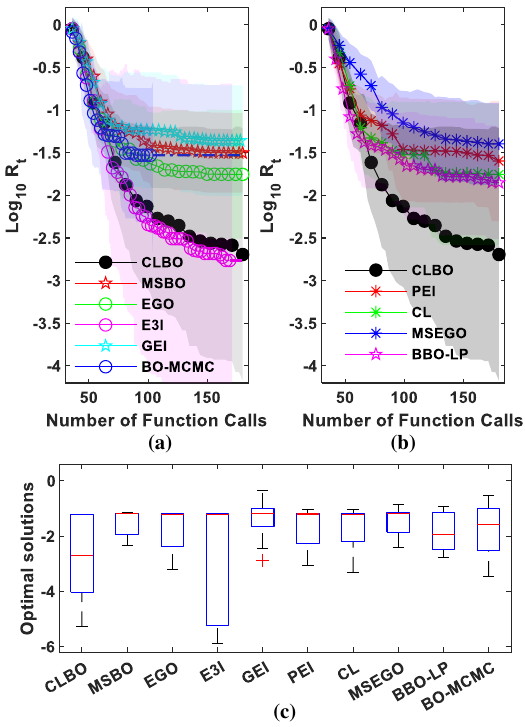}
	\caption{Test results of Hartman6 function, (a) comparison with sequential BO algorithms, (b) comparison with batch BO algorithms,
	(c) boxplot of the regrets of final optimal solutions}
	\label{fig:figure2-ai}
\end{figure}

\begin{figure}[t!]
	\centering
	\includegraphics[width=3in]{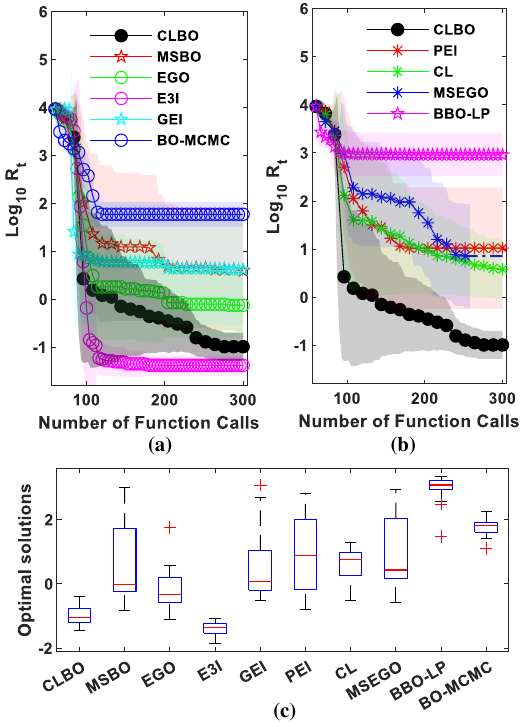}
	\caption{Test results of 10-dimensional Trid function, (a) comparison with sequential BO algorithms, (b) comparison with batch BO algorithms,
	(c) boxplot of the regrets of final optimal solutions}
	\label{fig:figure2-ai}
\end{figure}

\begin{table*}[htbp]
\centering
\caption{The averaged CPU time of compared algorithms when testing the five benchmark functions}
\renewcommand\arraystretch{1.2}
\begin{tabular}{clccccc}
 \toprule
Time (s) & Michalewicz & Rastrigin &Ackley& Hartman6  & Trid  \\
\hline
     CLBO 
     & $274.4$ 
     & $247.1$ 
     & $236.1$
     & $228.0$ 
     & $888.7$ \\
     MSBO 
     & $100.4$ 
     & $88.2$ 
     & $81.1$
     & $129.9$  
     & $321.9$ \\
     EGO 
     & $150.8$ 
     & $128.9$ 
     & $116.1$
     & $110.2$ 
     & $402.0$ \\

     E3I 
     & $2511.5$
     & $2618.0$ 
     & $2579.4$
     & $1731.6$
     & $4595.9$ \\
     
     GEI
     & $127.0$
     & $126.6$ 
     & $108.6$
     & $171.6$ 
     & $383.2$ \\    
     
     BO-MCMC
     & $6207.7$ 
     & $8681.0$ 
     & $4119.0$
     & $3426.3$ 
     & $16511.2$ \\
     
     PEI
     & $112.1$
     & $95.3$ 
     & $91.9$
     & $126.9$ 
     & $291.8$ \\
     
     CL
     & $164.1$ 
     & $149.9$ 
     & $125.4$
     & $124.1$ 
     & $480.4$ \\
     
     MSEGO
     & $693.5$ 
     & $653.1$ 
     & $727.1$
     & $1088.7$ 
     & $4505.9$ \\

     BBO-LP
     & $219.6$ 
     & $252.2$ 
     & $192.5$
     & $170.8$ 
     & $1030.2$ \\
\hline
\end{tabular}  
\end{table*}
\par

\begin{figure}[t!]
	\centering
	\includegraphics[width=3in]{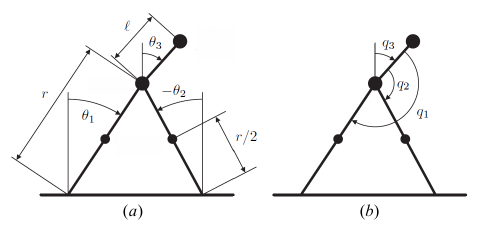}
	\caption{Schematic of a three-link bipedal robot~\cite{westervelt2018feedback}}
	\label{fig:figure2-ai}
\end{figure}

\begin{figure}[t!]
	\centering
	\includegraphics[width=2.5in]{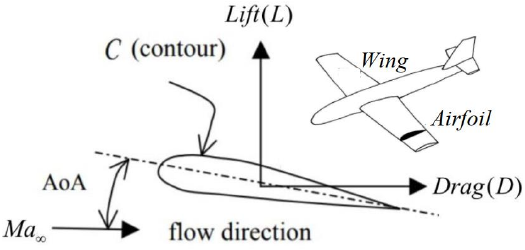}
	\caption{Schematic of airfoil contour and the forces acting on it~\cite{2021Guo}}
	\label{fig:figure2-ai}
\end{figure}

\subsection{Tests on Learning Problems}
To further show the effectiveness of CLBO, we also test it on engineering benchmarks, such as the control of robotics systems and airfoil optimization~\cite{chiu2020airfoil},~\cite{wang2018hierarchical}.
Note that simulation failures can be encountered when evaluating the samples for robotic systems and airfoil aerodynamic performance, which can get the algorithm stop early.
Hence, we propose to evaluate a random sample instead to restart the optimization process once encountered, which needs code modifications in related BO algorithms.
However, it may be not be easy for BBO-LP which implemented on a free-available software GPyOpt. On the other hand, as a contrast of CLBO, the poor performance of MSBO and the reason behind it have been discussed in the numerical tests.
Hence, BBO-LP, BO-MCMC and MSBO are not tested for engineering benchmark problems.
The detailed descriptions of the three selected engineering benchmarks are given as follows.

\begin{figure}[t!]
	\centering
	\includegraphics[width=3in]{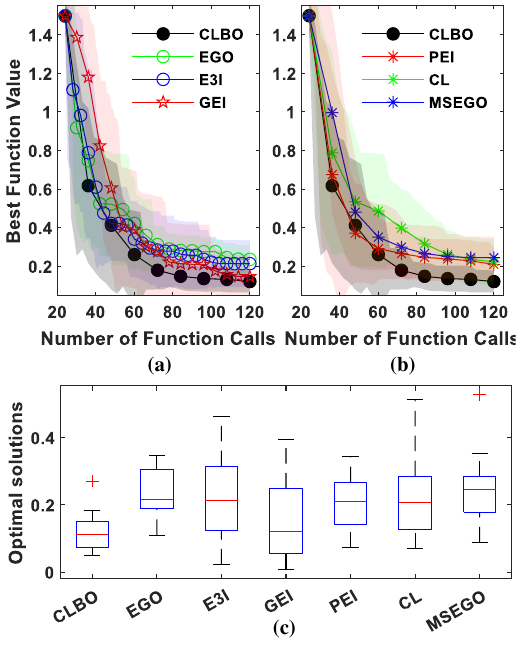}
	\caption{Test results of active learning for robot pushing, (a) comparison with sequential BO algorithms, (b) comparison with batch BO algorithms,
	(c) boxplot of the regrets of final optimal solutions}
	\label{fig:figure2-ai}
\end{figure}

\begin{figure}[t!]
	\centering
	\includegraphics[width=3in]{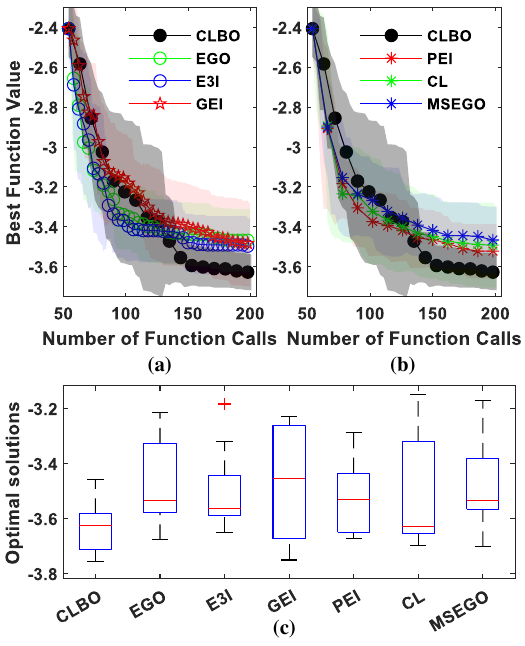}
	\caption{Test results of a planar bipedal walker reinforcement learning task, (a) comparison with sequential BO algorithms, (b) comparison with batch BO algorithms,
	(c) boxplot of the regrets of final optimal solutions}
	\label{fig:figure2-ai}
\end{figure}

\begin{figure}[t!]
	\centering
	\includegraphics[width=3in]{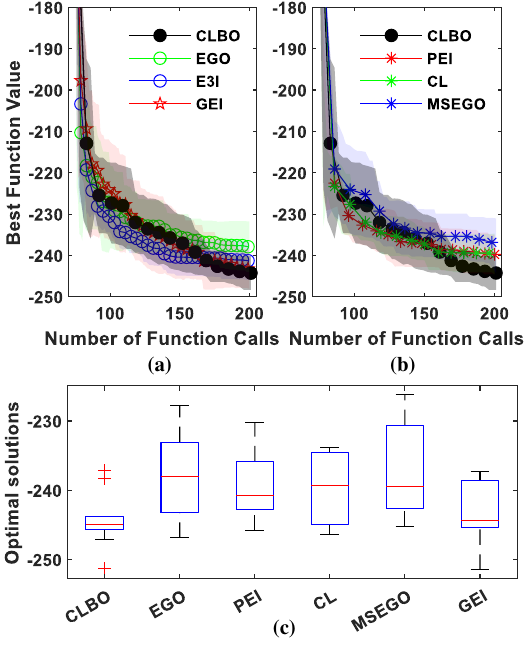}
	\caption{Test results of airfoil aerodynamic optimization, (a) comparison with sequential BO algorithms, (b) comparison with batch BO algorithms,
	(c) boxplot of the regrets of final optimal solutions}
	\label{fig:figure2-ai}
\end{figure}


\subsubsection{Active Learning for Robot Pushing} 
This task deals with the active learning for a pre-image learning problem for pushing~\cite{wang2017max}. The input is the pushing action of a robot, and the output is the distance of the pushed object to the expected location. The optimization objective is to find a good pre-image for pushing the object to the designated location. Four variables including the robot location and angle as (${r_x},{r_y},{r_\theta }$) and the pushing duration ($t$) are considered for the optimization of this task. 

\subsubsection{Reinforcement learning control of a planar bipedal walker} 
This problem considers the walking pattern control of a three-link bipedal walker in~\cite{westervelt2018feedback}, which is a reinforcement learning problem. Specifically, the controller modulates the walking pattern of a planar bipedal robot. The goal is to make the robot walking in a fast upright pattern, and the reward is the walking speed with a penalty for not maintaining the upright position. The controller has nine parameters as shown in Fig.7, where $r$ is the length of the leg, $l$ is the distance from the hips to the center of mass of the torso, $\left( {{\theta _{\rm{1}}},{\theta _{\rm{2}}},{\theta _3}} \right)$ are the absolute orientations of the various links, $({q_1},{q_2})$ are the relative angles and ${q_3}$ is the absolute orientation of the torso. In addition, there is one more parameter as $d{\theta _1}$, which is the input velocity of the stance leg. 

\begin{figure*}[t!]
	\centering
	\includegraphics[width=5in]{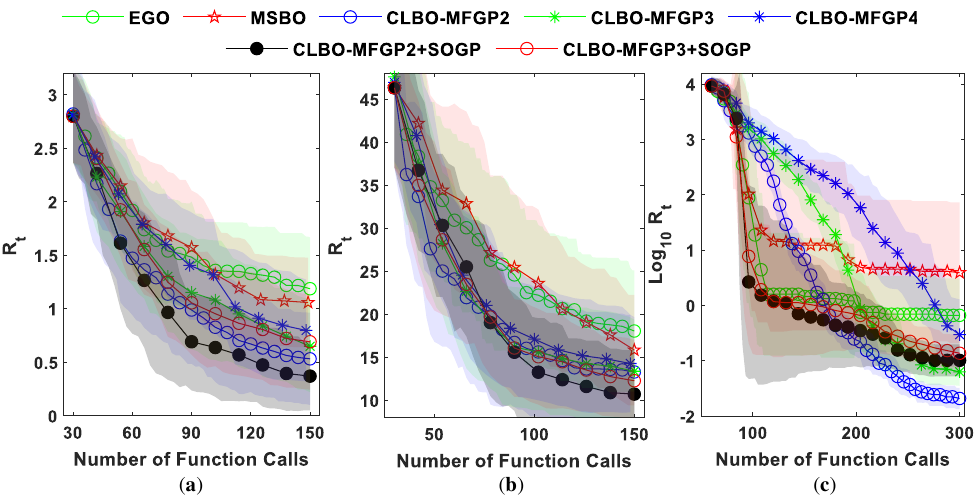}
	\caption{Sensitivity analysis of GP models by testing on representative benchmark functions, (a) Michalewicz function, (b) Rastrigin function, (c) Trid function}
	\label{fig:figure2-ai}
\end{figure*}
\subsubsection{The airfoil design problem} 
This problem considers the airfoil shape optimization in order to achieve excellent aerodynamic performance for aircraft wing.
As shown in Fig.8, the forces acting on the airfoil can be decomposed into two components as the drag ($D$) and the lift ($L$). 
The drag and lift and their ratio $L/D$ as well are the functions of the airfoil contour.
In the meantime, the airfoil contours can be in various shapes according to different working conditions determined by the aerodynamic parameters such as the Mach number ($Ma_{\infty }$), the Reynolds number ($Re$) and the angle of attack ($AoA$). 
In this study, we follow the settings in~\cite{2021Guo} for the optimization of a low-speed airfoil with 13 design variables.
In particular, the design objective is to maximize the lift to drag ratio, i.e.,\begin{small}$L/D$\end{small}, with the operating conditions set as \begin{small}$Re = 1.8 \times 10^6$\end{small}, \begin{small}$Ma$\end{small} = 0.01 and \begin{small}$AoA$\end{small} = 0 deg.
The XFOIL~\cite{drela1989xfoil} is used to calculate the objective performance $L/D$.
The detailed settings of 13 design variables for airfoil optimization can refer to~\cite{2021Guo}.

Figures 9 to 11 show the optimization results of engineering benchmark problems over 10 runs.
Similar to the results shown in the numerical tests, CLBO achieves the best solutions within budget.
For the 4-dimensional optimization problem as active learning for robot pushing, CLBO achieves better solutions almost from the start of the optimization process. 
However, when optimizing the 9-dimensional bipedal walker and 13-dimensional airfoil design problems, CLBO progresses slower at the early stage.
The slower convergence rate of CLBO may attribute to the poor prediction accuracy built with a small portion of subset samples at the early stage.
However, the situation is changed by imposing agreement constraint with sequentially added new samples, as seen by the faster convergence rate of CLBO at the later stage.
In contrast, MSEGO, which also uses multiple surrogate models to guide the optimization process, obtains much worse results at the end of the optimization process.

More importantly, as shown in the boxplots in Figs.9(c) to 11(c), the variation of final solutions of CLBO is the smallest among the compared algorithms, and the medians of CLBO are also better. 
The above can attribute to the exploiting of model agreement on unlabeled information. On one hand, the convergence of BO is highly related to the surrogate prediction accuracy in the vicinity of optimal solutions.
On the other hand, the agreement constraint on curve bumpiness helps to reduce the sample complexity of the effective hypothesis space~\cite{leskes2008value}.
Then, with the same number of subset samples, the agreement constrained GP predictions can be expected to achieve better prediction accuracy in neighborhood of optimal solutions(see Eq.(2)).
Additionally, the single-output GP trained with the full samples also help to increase the overall surrogate accuracy, by increasing the model diversity without scarifying the individual model error in BO process.
With the above, CLBO outperforms both the sequential and batch BO algorithms.



\subsection{Sensitivity Analysis on the Selection of GP Models}

To address the suboptimal problem of BO, we propose to exploit both model diversity and agreement on unlabeled information to improve the overall surrogate accuracy in BO process. In doing so, the agreement-constraint MFGP is combined with SOGP to build CLBO, their influences on algorithm performance are examined in this subsection.

Specifically, three versions of CLBO that use MFGP alone are built to inspect the effect of agreement-constrained modeling, which are labeled as CLBO-MFGP2, CLBO-MFGP3 and CLBO-MFGP4, respectively. 
In particular, CLBO-MFGP2 means two GP predictions are built with bootstrapping based sub-sampling (see Algorithm 2) in CLBO, similar treatment for CLBO-MFGP3 and CLBO-MFGP4.
Additionally, two more versions of CLBO that combines MFGP with SOGP are also tested, denoted by CLBO-MFGP2+SOGP and CLBO-MFGP3+SOGP, respectively. 
As contrast, the results of EGO and MSBO are also presented, as shown in Fig.12.

Among the five versions of CLBO, CLBO-MFGP2+SOGP achieves the best solutions when optimizing the Michalewicz and Rastrigin functions, and CLBO-MFGP2 performs the best on Trid. 
More interestingly, the convergence rates of CLBO-MFGP2+SOGP and CLBO-MFGP2 at the early stage of the optimization process are opposite to their final results. 
The reason behind can be explained as follows.
For the multi-peak functions like Michalewicz and Rastrigin, the agreement-constrained MFGP predictions can have reasonable accuracy in promising areas where relatively better solutions may locate. 
Then, \emph{similar to the local exploitation, the agreement on curve bumpiness confines the searching space of MFGP predictions to smaller yet promising areas, and hence CLBO-MFGP2 convergences more rapidly at the early stage of optimization process}.
Nevertheless, the accuracy of MFGP predictions may be poor in the region where even better solutions locate, but the SOGP trained with full samples can complement the above inaccuracy of MFGP predictions. As a consequence, CLBO-MFGP2-SOGP finally achieves better solutions.

On the other hand, for the high-dimensional convex functions like Trid, the accuracy of MFGP predictions trained with subset samples can be too poor to capture the global trend of the high-dimensional objective landscape at the early stage, and hence CLBO-MFGP2 progresses slowly at this stage.
As contrast, the SOGP trained with full samples can have reasonable accuracy over space, and hence CLBO-MFGP2-SOGP converges much faster at the early the stage when optimizing Trid.
Nevertheless, at later stage when MFGP predictions have sufficient accuracy to capture the convex function trend, the agreement on curve bumpiness makes CLBO-MFGP2 focus exploiting promising areas. And then, CLBO-MFGP2 progresses much faster and finally achieves even better solutions for this convex function.  

Also note that, the performance of CLBO becomes worse with the increase of the number of agreement-constrained MFGP predictions. 
The reason can be explained as follows.
As pointed out by~\cite{leskes2008value}, \emph{the extent of performance improvement by enforcing model agreement on unlabeled information depends on the diversity of the individual models}.
In particular for our study, increasing MFGP predictions can reduce the differences between MFGP and SOGP predictions. 
As a consequence, the overall surrogate accuracy in promising areas can become poorer (see Eq.(1)), resulting in slower convergence rate and worse final solutions. 




\section{Conclusions}

In this paper, we propose a co-learning Bayesian optimization framework (CLBO),  which exploits both diversity and agreement of multiple GP models on unlabeled information to improve the surrogate accuracy in BO process, and therefore achieving more efficient global optimization. 
Through tests on both numerical toy problems and various engineering benchmarks, 
it shows that both the agreement constraint and the diversity generated by multiple models contribute to the success of CLBO. 
Meanwhile, as CLBO can be run in parallel, the encouraging results of CLBO may also provide clues to resolve the exploitation-exploration challenge in batch BO. 
In future, we will further examine the advantages of CLBO with acquisition functions other than EI and parallel strategies that previously proposed under a single GP.
More importantly, we will explore more efficient ways to use the model agreements/disagreements on unlabeled information to further boost the BO performance.


\bibliography{References}

\begin{thebibliography}{10}
\providecommand{\url}[1]{#1}
\csname url@samestyle\endcsname
\providecommand{\newblock}{\relax}
\providecommand{\bibinfo}[2]{#2}
\providecommand{\BIBentrySTDinterwordspacing}{\spaceskip=0pt\relax}
\providecommand{\BIBentryALTinterwordstretchfactor}{4}
\providecommand{\BIBentryALTinterwordspacing}{\spaceskip=\fontdimen2\font plus
\BIBentryALTinterwordstretchfactor\fontdimen3\font minus \fontdimen4\font\relax}
\providecommand{\BIBforeignlanguage}[2]{{%
\expandafter\ifx\csname l@#1\endcsname\relax
\typeout{** WARNING: IEEEtran.bst: No hyphenation pattern has been}%
\typeout{** loaded for the language `#1'. Using the pattern for}%
\typeout{** the default language instead.}%
\else
\language=\csname l@#1\endcsname
\fi
#2}}
\providecommand{\BIBdecl}{\relax}
\BIBdecl

\bibitem{martinez2018funneled}
R.~Martinez-Cantin, ``Funneled bayesian optimization for design, tuning and control of autonomous systems,'' \emph{IEEE transactions on cybernetics}, no.~99, pp. 1--12, 2018.

\bibitem{2021Guo}
Z.~Guo, Y.~Ong, and H.~Liu, ``Calibrated and recalibrated expected improvements for bayesian optimization,'' \emph{Structural and Multidisciplinary Optimization}, 2021.

\bibitem{song2016research}
L.~Song, Z.~Guo, J.~Li, and Z.~Feng, ``Research on metamodel-based global design optimization and data mining methods,'' \emph{Journal of Engineering for Gas Turbines and Power}, vol. 138, no.~9, 2016.

\bibitem{martinez2017feeling}
U.~Martinez-Hernandez, T.~J. Dodd, and T.~J. Prescott, ``Feeling the shape: Active exploration behaviors for object recognition with a robotic hand,'' \emph{IEEE Transactions on Systems, Man, and Cybernetics: Systems}, vol.~48, no.~12, pp. 2339--2348, 2017.

\bibitem{swersky2013multi}
K.~Swersky, J.~Snoek, and R.~P. Adams, ``Multi-task bayesian optimization,'' in \emph{Advances in neural information processing systems}, 2013, pp. 2004--2012.

\bibitem{rasmussen2010gaussian}
C.~E. Rasmussen and H.~Nickisch, ``Gaussian processes for machine learning (gpml) toolbox,'' \emph{Journal of machine learning research}, vol.~11, no. Nov, pp. 3011--3015, 2010.

\bibitem{jones1998efficient}
D.~R. Jones, M.~Schonlau, and W.~J. Welch, ``Efficient global optimization of expensive black-box functions,'' \emph{Journal of Global optimization}, vol.~13, no.~4, pp. 455--492, 1998.

\bibitem{couckuyt2014fast}
I.~Couckuyt, D.~Deschrijver, and T.~Dhaene, ``Fast calculation of multiobjective probability of improvement and expected improvement criteria for pareto optimization,'' \emph{Journal of Global Optimization}, vol.~60, no.~3, pp. 575--594, 2014.

\bibitem{hennig2012entropy}
P.~Hennig and C.~J. Schuler, ``Entropy search for information-efficient global optimization,'' \emph{Journal of Machine Learning Research}, vol.~13, no. Jun, pp. 1809--1837, 2012.

\bibitem{wu2016parallel}
J.~Wu and P.~Frazier, ``The parallel knowledge gradient method for batch bayesian optimization,'' in \emph{Advances in Neural Information Processing Systems}, 2016, pp. 3126--3134.

\bibitem{stuckman1992comparison}
B.~E. Stuckman and E.~E. Easom, ``A comparison of bayesian/sampling global optimization techniques,'' \emph{IEEE Transactions on Systems, Man, and Cybernetics}, vol.~22, no.~5, pp. 1024--1032, 1992.

\bibitem{huang2006global}
D.~Huang, T.~T. Allen, W.~I. Notz, and N.~Zeng, ``Global optimization of stochastic black-box systems via sequential kriging meta-models,'' \emph{Journal of global optimization}, vol.~34, no.~3, pp. 441--466, 2006.

\bibitem{bull2011convergence}
A.~D. Bull, ``Convergence rates of efficient global optimization algorithms,'' \emph{Journal of Machine Learning Research}, vol.~12, no. Oct, pp. 2879--2904, 2011.

\bibitem{qin2017improving}
C.~Qin, D.~Klabjan, and D.~Russo, ``Improving the expected improvement algorithm,'' in \emph{Advances in Neural Information Processing Systems}, 2017, pp. 5381--5391.

\bibitem{2002Flexibility}
M.~J. Sasena, ``Flexibility and efficiency enhancements for constrained global design optimization with kriging approximations,'' Ph.D. dissertation, University of Michigan., 2002.

\bibitem{shahriari2015taking}
B.~Shahriari, K.~Swersky, Z.~Wang, R.~P. Adams, and N.~De~Freitas, ``Taking the human out of the loop: A review of bayesian optimization,'' \emph{Proceedings of the IEEE}, vol. 104, no.~1, pp. 148--175, 2015.

\bibitem{joy2019flexible}
T.~T. Joy, S.~Rana, S.~Gupta, and S.~Venkatesh, ``A flexible transfer learning framework for bayesian optimization with convergence guarantee,'' \emph{Expert Systems with Applications}, vol. 115, pp. 656--672, 2019.

\bibitem{min2020generalizing}
A.~T.~W. Min, A.~Gupta, and Y.-S. Ong, ``Generalizing transfer bayesian optimization to source-target heterogeneity,'' \emph{IEEE Transactions on Automation Science and Engineering}, 2020.

\bibitem{kandasamy2017multi}
K.~Kandasamy, G.~Dasarathy, J.~Schneider, and B.~P{\'o}czos, ``Multi-fidelity bayesian optimisation with continuous approximations,'' in \emph{Proceedings of the 34th International Conference on Machine Learning-Volume 70}.\hskip 1em plus 0.5em minus 0.4em\relax JMLR. org, 2017, pp. 1799--1808.

\bibitem{guo2018analysis}
Z.~Guo, L.~Song, C.~Park, J.~Li, and R.~T. Haftka, ``Analysis of dataset selection for multi-fidelity surrogates for a turbine problem,'' \emph{Structural and Multidisciplinary Optimization}, vol.~57, no.~6, pp. 2127--2142, 2018.

\bibitem{2021Parallel}
Z.~Guo, Q.~Wang, L.~Song, and J.~Li, ``Parallel multi-fidelity expected improvement method for efficient global optimization,'' \emph{Structural and Multidisciplinary Optimization}, 2021.

\bibitem{gupta2017insights}
A.~Gupta, Y.-S. Ong, and L.~Feng, ``Insights on transfer optimization: Because experience is the best teacher,'' \emph{IEEE Transactions on Emerging Topics in Computational Intelligence}, vol.~2, no.~1, pp. 51--64, 2017.

\bibitem{zhang2015multimodel}
Q.~Zhang, C.~Zhou, N.~Xiong, Y.~Qin, X.~Li, and S.~Huang, ``Multimodel-based incident prediction and risk assessment in dynamic cybersecurity protection for industrial control systems,'' \emph{IEEE Transactions on Systems, Man, and Cybernetics: Systems}, vol.~46, no.~10, pp. 1429--1444, 2015.

\bibitem{min2017multiproblem}
A.~T.~W. Min, Y.-S. Ong, A.~Gupta, and C.-K. Goh, ``Multiproblem surrogates: transfer evolutionary multiobjective optimization of computationally expensive problems,'' \emph{IEEE Transactions on Evolutionary Computation}, vol.~23, no.~1, pp. 15--28, 2017.

\bibitem{zhou2006combining}
Z.~Zhou, Y.~S. Ong, P.~B. Nair, A.~J. Keane, and K.~Y. Lum, ``Combining global and local surrogate models to accelerate evolutionary optimization,'' \emph{IEEE Transactions on Systems, Man, and Cybernetics, Part C (Applications and Reviews)}, vol.~37, no.~1, pp. 66--76, 2006.

\bibitem{2013Efficient}
F.~A.~C. Viana, R.~T. Haftka, and L.~T. Watson, ``Efficient global optimization algorithm assisted by multiple surrogate techniques,'' \emph{Journal of Global Optimization}, vol.~56, no.~2, pp. 669--689, 2013.

\bibitem{he2019contextual}
T.~He, Y.~Liu, T.~H. Ko, K.~C. Chan, and Y.-S. Ong, ``Contextual correlation preserving multiview featured graph clustering,'' \emph{IEEE transactions on cybernetics}, 2019.

\bibitem{xu2013survey}
C.~Xu, D.~Tao, and C.~Xu, ``A survey on multi-view learning,'' \emph{arXiv preprint arXiv:1304.5634}, 2013.

\bibitem{zhou2009semi}
Z.-H. Zhou, ``When semi-supervised learning meets ensemble learning,'' in \emph{International Workshop on Multiple Classifier Systems}.\hskip 1em plus 0.5em minus 0.4em\relax Springer, 2009, pp. 529--538.

\bibitem{krogh1995neural}
A.~Krogh and J.~Vedelsby, ``Neural network ensembles, cross validation, and active learning,'' in \emph{Advances in neural information processing systems}, 1995, pp. 231--238.

\bibitem{Luca2015Local}
Luca, Oneto, Alessandro, Ghio, Sandro, Ridella, Davide, and Anguita, ``Local rademacher complexity: Sharper risk bounds with and without unlabeled samples,'' \emph{Neural Networks}, 2015.

\bibitem{leskes2008value}
B.~Leskes and L.~Torenvliet, ``The value of agreement a new boosting algorithm,'' \emph{Journal of Computer and System Sciences}, vol.~74, no.~4, pp. 557--586, 2008.

\bibitem{farquhar2006two}
J.~Farquhar, D.~Hardoon, H.~Meng, J.~S. Shawe-Taylor, and S.~Szedmak, ``Two view learning: Svm-2k, theory and practice,'' in \emph{Advances in neural information processing systems}, 2006, pp. 355--362.

\bibitem{blum1998combining}
A.~Blum and T.~Mitchell, ``Combining labeled and unlabeled data with co-training,'' in \emph{Proceedings of the eleventh annual conference on Computational learning theory}, 1998, pp. 92--100.

\bibitem{navaratnam2007joint}
R.~Navaratnam, A.~W. Fitzgibbon, and R.~Cipolla, ``The joint manifold model for semi-supervised multi-valued regression,'' in \emph{2007 IEEE 11th International Conference on Computer Vision}.\hskip 1em plus 0.5em minus 0.4em\relax IEEE, 2007, pp. 1--8.

\bibitem{zhou2004democratic}
Y.~Zhou and S.~Goldman, ``Democratic co-learning,'' in \emph{16th IEEE International Conference on Tools with Artificial Intelligence}.\hskip 1em plus 0.5em minus 0.4em\relax IEEE, 2004, pp. 594--602.

\bibitem{muslea2006active}
I.~Muslea, S.~Minton, and C.~A. Knoblock, ``Active learning with multiple views,'' \emph{Journal of Artificial Intelligence Research}, vol.~27, pp. 203--233, 2006.

\bibitem{bonilla2008multi}
E.~V. Bonilla, K.~M. Chai, and C.~Williams, ``Multi-task gaussian process prediction,'' in \emph{Advances in neural information processing systems}, 2008, pp. 153--160.

\bibitem{2021Revisiting}
R.~L. Riche and V.~Picheny, ``Revisiting bayesian optimization in the light of the coco benchmark,'' 2021.

\bibitem{ginsbourger2010kriging}
D.~Ginsbourger, R.~Le~Riche, and L.~Carraro, ``Kriging is well-suited to parallelize optimization,'' in \emph{Computational intelligence in expensive optimization problems}.\hskip 1em plus 0.5em minus 0.4em\relax Springer, 2010, pp. 131--162.

\bibitem{J2015Batch}
J.~González, Z.~Dai, P.~Hennig, and N.~D. Lawrence, ``Batch bayesian optimization via local penalization,'' \emph{statistics}, 2015.

\bibitem{1998Global}
M.~Schonlau and W.~D.~R. Jones, \emph{Global versus local search in constrained optimization of computer models}.\hskip 1em plus 0.5em minus 0.4em\relax Institute of Mathematical Statistics, 1998.

\bibitem{ponweiser2008clustered}
W.~Ponweiser, T.~Wagner, and M.~Vincze, ``Clustered multiple generalized expected improvement: A novel infill sampling criterion for surrogate models,'' in \emph{2008 IEEE Congress on Evolutionary Computation (IEEE World Congress on Computational Intelligence)}.\hskip 1em plus 0.5em minus 0.4em\relax IEEE, 2008.

\bibitem{berk2018exploration}
J.~Berk, V.~Nguyen, S.~Gupta, S.~Rana, and S.~Venkatesh, ``Exploration enhanced expected improvement for bayesian optimization,'' in \emph{Joint European Conference on Machine Learning and Knowledge Discovery in Databases}.\hskip 1em plus 0.5em minus 0.4em\relax Springer, 2018.

\bibitem{liu2018remarks}
H.~Liu, J.~Cai, and Y.-S. Ong, ``Remarks on multi-output gaussian process regression,'' \emph{Knowledge-Based Systems}, vol. 144, pp. 102--121, 2018.

\bibitem{durichen2014multi}
R.~D{\"u}richen, M.~A. Pimentel, L.~Clifton, A.~Schweikard, and D.~A. Clifton, ``Multi-task gaussian process models for biomedical applications,'' in \emph{IEEE-EMBS International Conference on Biomedical and Health Informatics (BHI)}.\hskip 1em plus 0.5em minus 0.4em\relax IEEE, 2014, pp. 492--495.

\bibitem{pan2009survey}
S.~J. Pan and Q.~Yang, ``A survey on transfer learning,'' \emph{IEEE Transactions on knowledge and data engineering}, vol.~22, no.~10, pp. 1345--1359, 2009.

\bibitem{luo2018evolutionary}
J.~Luo, A.~Gupta, Y.-S. Ong, and Z.~Wang, ``Evolutionary optimization of expensive multiobjective problems with co-sub-pareto front gaussian process surrogates,'' \emph{IEEE transactions on cybernetics}, vol.~49, no.~5, pp. 1708--1721, 2018.

\bibitem{wei2017source}
P.~Wei, R.~Sagarna, Y.~Ke, Y.-S. Ong, and C.-K. Goh, ``Source-target similarity modelings for multi-source transfer gaussian process regression,'' in \emph{Proceedings of the 34th International Conference on Machine Learning-Volume 70}.\hskip 1em plus 0.5em minus 0.4em\relax JMLR. org, 2017, pp. 3722--3731.

\bibitem{da2018curbing}
B.~Da, A.~Gupta, and Y.-S. Ong, ``Curbing negative influences online for seamless transfer evolutionary optimization,'' \emph{IEEE transactions on cybernetics}, vol.~49, no.~12, pp. 4365--4378, 2018.

\bibitem{zhan2017pseudo}
D.~Zhan, J.~Qian, and Y.~Cheng, ``Pseudo expected improvement criterion for parallel ego algorithm,'' \emph{Journal of Global Optimization}, vol.~68, no.~3, pp. 641--662, 2017.

\bibitem{snoek2012practical}
J.~Snoek, H.~Larochelle, and R.~P. Adams, ``Practical bayesian optimization of machine learning algorithms,'' in \emph{Advances in neural information processing systems}, 2012, pp. 2951--2959.

\bibitem{gpyopt2016}
T.~G. authors, ``{GPyOpt}: A bayesian optimization framework in python,'' \url{http://github.com/SheffieldML/GPyOpt}, 2016.

\bibitem{wang2017max}
Z.~Wang and S.~Jegelka, ``Max-value entropy search for efficient bayesian optimization,'' in \emph{Proceedings of the 34th International Conference on Machine Learning-Volume 70}.\hskip 1em plus 0.5em minus 0.4em\relax JMLR. org, 2017, pp. 3627--3635.

\bibitem{simulationlib}
S.~Surjanovic and D.~Bingham, ``Virtual library of simulation experiments: Test functions and datasets,'' Retrieved June 10, 2021, from http://www.sfu.ca/~ssurjano.

\bibitem{zhang2018variable}
Y.~Zhang, Z.-H. Han, and K.-S. Zhang, ``Variable-fidelity expected improvement method for efficient global optimization of expensive functions,'' \emph{Structural and Multidisciplinary Optimization}, vol.~58, no.~4, pp. 1431--1451, 2018.

\bibitem{westervelt2018feedback}
E.~R. Westervelt, J.~W. Grizzle, C.~Chevallereau, J.~H. Choi, and B.~Morris, \emph{Feedback control of dynamic bipedal robot locomotion}.\hskip 1em plus 0.5em minus 0.4em\relax CRC press, 2018.

\bibitem{chiu2020airfoil}
K.~Chiu, M.~Fuge \emph{et~al.}, ``Airfoil design parameterization and optimization using b$\backslash$'ezier generative adversarial networks,'' \emph{arXiv preprint arXiv:2006.12496}, 2020.

\bibitem{wang2018hierarchical}
H.~Wang, J.~Doherty, and Y.~Jin, ``Hierarchical surrogate-assisted evolutionary multi-scenario airfoil shape optimization,'' in \emph{2018 IEEE Congress on Evolutionary Computation (CEC)}.\hskip 1em plus 0.5em minus 0.4em\relax IEEE, 2018, pp. 1--8.

\bibitem{drela1989xfoil}
M.~Drela, ``Xfoil: An analysis and design system for low reynolds number airfoils,'' in \emph{Low Reynolds number aerodynamics}.\hskip 1em plus 0.5em minus 0.4em\relax Springer, 1989.

\end{thebibliography}
\bibliographystyle{IEEEtran}
\end{document}